\documentclass[reqno,oneside,letterpaper]{amsart}
\usepackage{hyperref}
\usepackage{url}
\usepackage{comment}
\usepackage[style=numeric-comp,
       minnames=1,
       maxnames=99,
       maxcitenames=3,
       natbib=true,
       firstinits=true,
       backend=bibtex]{biblatex}

\usepackage[capitalize]{cleveref}  

\crefname{nlem}{Lemma}{Lemmas}
\crefname{nprop}{Proposition}{Propositions}
\crefname{ncor}{Corollary}{Corollaries}
\crefname{nthm}{Theorem}{Theorems}
\crefname{assumption}{Assumption}{Assumptions}

\usepackage{color}
\usepackage{jhh-misc2}

\boldshortcuts
\setshortness{0}

\allowdisplaybreaks[3]

\bibliography{sva}

\title%
[SVA for Some Hierarchical Bayesian Nonparametric Models]
{Detailed Derivations of Small-variance Asymptotics 
for some Hierarchical Bayesian Nonparametric Models}

\author[J.~H.~Huggins]{Jonathan H.~Huggins}
\address{Massachusetts Institute of Technology}
\urladdr{http://jhhuggins.org/}
\email{jhuggins@mit.edu}

\author[A.~Saeedi]{Ardavan Saeedi}
\address{Massachusetts Institute of Technology}
\email{ardavans@mit.edu}

\author[M.~J.~Johnson]{Matthew J.~Johnson}
\address{Harvard University}
\urladdr{http://www.mit.edu/~mattjj/}
\email{mattjj@csail.mit.edu}

\begin{document}

\begin{abstract}
In this note we provide detailed derivations of two versions of
small-variance asymptotics for hierarchical Dirichlet
process (HDP) mixture models and the HDP hidden 
Markov model (HDP-HMM, a.k.a.\ the infinite HMM).
We include derivations for the probabilities of
certain CRP and CRF partitions, which are of 
more general interest.
\end{abstract}

\maketitle

\section{Introduction}

Numerous flexible Bayesian nonparametric models
and associated inference algorithms have been
developed in recent years for solving problems such
as clustering and time series analysis.
However, simpler approaches such as $k$-means 
remain extremely popular 
due to their simplicity and scalability to the 
large-data setting. 

The $k$-means optimization problem can be viewed as the small-variance limit of MAP
inference in a $k$-component Gaussian mixture model.
That is, with observed data $\mcX = (x_{n})_{n=1}^{N}$, $x_{n} \in \reals^{D}$,
the Gaussian mixture model log joint density with means
$\mu_{1},\dots,\mu_{n} \in \reals^{D}$,
cluster assignments $\mcZ = (z_{n})_{n=1}^{N}$ with $z_n \in \{1,2,\ldots,K\}$,
and spherical variance $\sigma^{2}$ is
\[
\begin{split}
  \log p(\mu, \mcZ, \mcX) 
  &= \log p(\mu) p(\mcZ) - \frac{1}{2} ND \log 2 \pi \sigma^2 - \sum_{n=1}^N \frac{||x_n - \mu_{z_n}||^2}{2 \sigma^2} \\
  &= \beta \sum_{n=1}^N ||x_n - \mu_{z_n} ||^2 + o(\beta),
\end{split}
\]
where $\beta \defined \frac{1}{2 \sigma^2}$.
As $\sigma^2 \to 0$, or equivalently $\beta \to \infty$, the term
that is linear in $\beta$ dominates and the MAP problem becomes the $k$-means
problem in the sense that
\[
\begin{split}
  \lim_{\sigma^2 \to 0} \argmax_{\mcZ,\mu} \log p(\mu,\mcZ,\mcX)
  &= \lim_{\beta \to \infty} \argmin_{\mcZ,\mu} \beta \sum_n ||x_n - \mu_{z_n}||^2 + o(\beta) \\
  &= \argmin_{\mcZ,\mu} \sum_n ||x_n - \mu_{z_n} ||^2.
\end{split}
\]
Note that we have assumed the priors $p(\mcZ)$ and $p(\mu)$ are positive
and independent of $\sigma^{2}$.

Recently developed small-variance asymptotics (SVA)
\citep{Kulis:2012,Jiang:2012,Broderick:2013} methods
generalize the above derivation of $k$-means to other
Bayesian models, with nonparametric Bayesian models
such as those based on the Dirichlet process (DP) and 
the Indian buffet process being of particular interest. 
While obtaining $k$-means from the Gaussian mixture model
is straightforward, the SVA derivations for nonparametric
models can be quite subtle, especially for hierarchical models. 
Indeed, we are not aware of a reference with 
the derivations for many important DP and hierarchal DP
(HDP) probability expressions.
This note is meant to serve as a self-contained 
reference to some DP and HDP material of general 
interest as well as SVA derivations for HDP-based models.
In particular, we provide derivations for the 
HDP mixture model and the HDP-HMM (a.k.a.~the iHMM).

\subsection*{A Caution.}
This note is \emph{not} meant to serve as an 
introduction to SVA methods or Bayesian nonparametric
modeling tools. 
Thus, we assume the reader is familiar with the MAD-Bayes
approach to SVA \citep{Broderick:2013} and scaled
exponential family distributions \citep{Jiang:2012}.
We also assume basic familiarity with the Chinese
restaurant process (CRP) and Chinese restaurant franchise 
(CRF) representations of, respectively, the DP and the HDP
\citep{Aldous:1985,Teh:2006b}.

\section{Preliminaries}

\subsection{Notation}

For an arbitrary real-valued vector $\bv \in \reals^{D}$ 
and indices $1 \le i \le j \le D$, let 
$\bv_{i:j} \defined \tuple[v_{i},v_{i+1},\dots,v_{j}]$,
$\bar v_{j} \defined \sum_{i=1}^{j} v_{i}$, and
$v_{\cdot} = \bar v_{D}$;
we extend the range and dot notations in the obvious way 
to matrices and tensors. 
By convention $\prod_{\emptyset} = 1$ and 
$\sum_{\emptyset} = 0$.

\subsection{Dirichlet process, Chinese restaurants, and all that}

Recall that for a CRP with
concentration parameter $\kappa$, given
that there are $c$ observations, the 
probability of observing $L$ tables with 
counts $c_{1},\dots,c_{L}$ is:
\[
\Pr_{CRP}(\bc) 
&= \prod_{\ell=1}^{L}\frac{\kappa \cdot c_{\ell}!}{\prod_{i=\bar c_{\ell-1}}^{\bar c_{\ell}-1} (\kappa + i)} \\
&= \frac{\kappa^{L}\Gamma(\kappa)}{\Gamma(\kappa + c_{\cdot})}\prod_{\ell=1}^{L} c_{\ell}! \\
&= \frac{\kappa^{L-1}\Gamma(\kappa + 1)}{\Gamma(\kappa + c_{\cdot})}\prod_{\ell=1}^{L} c_{\ell}!, \label{eq:crp-prob}
\]
where we have used the exchangeability of the CRP. 

For a CRF with concentration parameters $\kappa$ and $\alpha$
and $c$ observations,
let $K$ be the number of tables in the top-level restaurant,
let $N$ be the number of franchises, let $t_{ij}$ be the 
number of tables in restaurant franchise $i$ serving dish $j$,
and let $c_{ijt}$ be the number of customers
at the $t$-th table serving dish $j$ in restaurant $i$. 
For the top-level restaurant, the ``customer'' counts are 
$(t_{i\cdot})_{i}$ and for $i$-th franchise, the customer
counts are $(c_{i\cdot j})_{j}$. 
Hence, repeatedly using \cref{eq:crp-prob}, we have
\[
\begin{split}
\lefteqn{\Pr_{CRF}(\bt, \bc \given \alpha, \kappa)}  \\
&= \frac{\kappa^{K-1}\Gamma(\kappa + 1)}{\Gamma(\kappa + t_{\cdot\cdot})}\prod_{j=1}^{K} t_{\cdot j}!
  \prod_{i=1}^{N}\left(\frac{\alpha^{t_{i\cdot} - 1}\Gamma(\alpha + 1)}{\Gamma(\alpha + c_{i\cdot\cdot})}\prod_{j=1}^{K}\prod_{t=1}^{t_{ij}}c_{ijt}!  \right).
\end{split}
\]
We can integrate over all possible seating arrangements for the 
customers in the franchises.
That is, if $C_{ij} = c_{ij\cdot}$ is the number of 
customers eating dish $j$ at restaurant $i$, then
\[
\Pr_{CRF}(\bt, \bC \given \alpha, \kappa) 
&= \frac{\kappa^{K-1}\Gamma(\kappa+1)}{\Gamma(\kappa + t_{\cdot\cdot})}\prod_{j=1}^{K} t_{\cdot j}!
  \prod_{i=1}^{N}\left(\frac{\alpha^{t_{i\cdot}-1}\Gamma(\alpha+1)}{\Gamma(\alpha + C_{i\cdot})}\prod_{j=1}^{K}\begin{bmatrix} C_{ij} \\ t_{ij} \end{bmatrix}  \right), \label{eq:crf-prob}
\]
where $\begin{bmatrix} C_{ij} \\ t_{ij} \end{bmatrix}$ is an unsigned Stirling
number of the first kind \citep{Antoniak:1974}, which can be interpreted as
the number of ways to seat $C_{ij}$ customers 
at $t_{ij}$ tables such that each table has at least one customer
(more formally, the number of permutations of
$C_{ij}$ objects with $t_{ij}$ disjoint cycles). 

\section{SVA for HDP mixture models}
\label{sec:SVAforHDP}
The generative model for the HDP \citep{Teh:2006b} with $N$ groups and $J_i$ Gaussian observations in group $i$ is:
\[
\bbeta &\dist \distNamed{GEM}(\gamma) \\
\bpi_{i} &\dist \distDP(\alpha\bbeta), & i &\ge 1 \\
\mu_{k} &\dist \distNorm(0,\sigma_{0}^{2}), & k &\ge 1 \\
z_{ij}  &\dist \distMulti(\bpi_{i}), & j, i &\ge 1 \\
y_{ij} \given z_{ij} &\dist \distNorm(\mu_{z_{ij}}, \sigma^{2}), & j ,i &\ge 1,
\]
Here $\distNamed{GEM}(\gamma)$ is the stick-breaking prior
with concentration parameter $\gamma$ \citep{Sethuraman:1994} and $y_{ij}$ 
is the $j^{th}$ observation in the $i^{th}$ group. 
Let $\mcZ_i \defined (z_{ij})_{j=1}^{J_i}$, $\mcY_i \defined (y_{ij})_{j=1}^{J_i}$, 
and $K \defined \max_{ij}z_{ij}$.
For the joint density of the HDP we have: 
 \[
\begin{split}
\lefteqn{\Pr(\mcZ_{1:N}, \mcY_{1:N}, \bmu_{1:N}, \bpi_{1:N,1:K}, \bbeta_{1:K} \given \sigma_{0}, \sigma, \alpha, \gamma)} \\
& = \Pr(\mcZ_{1:N}, \bbeta_{1:K}, \bpi_{1:N,1:K} \given \alpha, \gamma) 
\prod_{k=1}^{K} \prod_{ij : z_{ij} = k} \distNorm(y_{ij}\given \mu_{k}, \sigma^{2})
\prod_{k=1}^{K} \distNorm(\mu_{k}\given 0, \sigma_{0}^{2}),
\end{split}
\]
where
\[
\begin{split}
\lefteqn{\Pr(\mcZ_{1:N}, \bbeta_{1:K}, \bpi_{1:N,1:K} \given \alpha, \gamma)} \\
&\defined 
\distNamed{GEM}(\bbeta_{1:K}; \gamma)
\prod_{i=1}^{N} \distDP(\bpi_{i,1:K}; \alpha\bbeta_{1:K})
\prod_{i=1}^{N} \prod_{j=1}^{J_i} \distMulti(z_{ij}; \bpi_{i,1:K}). \label{eq:hdp}
\end{split}
\]
Integrating out $\bbeta_{1:K}$ and $\bpi_{1:N}$, we obtain the CRF representation in \cref{eq:crf-prob}:
\[
\Pr(\bt, \mcZ_{1:N} \given \alpha, \gamma) 
&= \frac{\gamma^{K-1}\Gamma(\gamma+1)}{\Gamma(\gamma + t_{\cdot\cdot})}\prod_{j=1}^{K} t_{\cdot j}!
  \prod_{i=1}^{N}\left(\frac{\alpha^{t_{i\cdot}-1}\Gamma(\alpha+1)}{\Gamma(\alpha + z_{i\cdot})}\prod_{j=1}^{K}\begin{bmatrix} z_{ij} \\ t_{ij} \end{bmatrix}  \right), \label{eq:hdp-crf}
\]
As in the introduction, define $\beta = \frac{1}{2\sigma^2}$, 
so we are interested in the limit $\beta \to \infty$
(i.e., $\sigma^{2} \to 0$).
To maintain the effects of the hyperparameters in the
small-variance limit, we set $\gamma = \exp(-\lambda_{1}\beta)$
and $\alpha = \exp(-\lambda_{2}\beta)$, where
$\lambda_1 > 0$ and $\lambda_2 > 0$ are free parameters.
Taking the logarithm of \cref{eq:crf-prob}, we obtain
\[
\log\Pr(\bt, \mcZ_{1:N} \given \alpha, \gamma) 
&= (K - 1) \log \gamma + \log \frac{\Gamma(\gamma+1)}{\Gamma(\gamma + t_{\cdot\cdot})}+\sum_{j=1}^K \log t_{\cdot j}! \label{eq:log-hdp-crf-1} \\
&\phantom{=~} + \sum_{i=1}^N \left\{(t_{i \cdot} - 1)\log \alpha + \log \frac{\Gamma(\alpha + 1)}{\Gamma(\alpha + z_{i\cdot})} + \sum^K_{j=1}\log \begin{bmatrix} z_{ij} \\ t_{ij} \end{bmatrix} \right\} \nonumber \\
&= -\beta \lambda_{1} (K-1) + O(1) -
\sum_{i=1}^{N}\{  \beta \lambda_{2}(t_{i\cdot} - 1) + O(1)\}. \label{eq:log-hdp-crf-2}
\]
If $m_{i} \defined \#\{ z_{ij} \}_{i=1}^{J_{i}}$ is the number of distinct 
indices in $\mcZ_{i}$, then 
\[
\max_{\bt \given \bt \sim \mcZ_{1:N}} - \lambda_{1} (K-1) - \lambda_{2}\sum_{i=1}^{K} (t_{i\cdot} - 1)
= - \lambda_{1} (K-1) - \lambda_{2}\sum_{i=1}^{K} (m_{i} - 1),
\]
where $\bt \sim \mcZ_{1:T}$ denotes that $\bt$ is consistent 
with $\mcZ_{1:T}$.
Hence, after setting the variance of $\mu_{k}$ to be 
$\sigma^2_0 = \sigma^2 / \lambda_3$, where $\lambda_{3} \ge 0$
is a free parameter, the SVA objective function for the HDP 
mixture model is
\[
\min_{K,\mcZ,\bmu} \left\{
  \sum_{k=1}^{K} \sum_{ij : z_{ij} = k} \|y_{ij} - \mu_{k}\|^{2} + \lambda_{1} (K-1) 
+ \lambda_{2}\sum_{i=1}^{N} (m_{i} - 1) 
+ \lambda_{3}\sum_{k=1}^{K} \|\mu_{k}\|^{2} \right\}.  \label{eq:sva-hdp}
\]

This cost function is in fact the $k$-means objective function with some additional penalty terms. 
The second and third terms in \cref{eq:sva-hdp} penalize the number of global and local clusters, 
respectively. 
The final term introduces the additional cost for the prior over the cluster means.  

\section{SVA for the HDP-HMM}

The HDP-HMM generative model with Gaussian observations
is \citep{VanGael:2008}:
\[
\bbeta &\dist \distNamed{GEM}(\gamma) \\
\bpi_{k} &\dist \distDP(\alpha\bbeta), & k &\ge 1 \\
\mu_{k} &\dist \distNorm(0,\sigma_{0}^{2}), & k &\ge 1 \\
z_{t} \given z_{t-1} &\dist \distMulti(\bpi_{z_{t-1}}), & t &\ge 2 \\
y_{t} \given z_{t} &\dist \distNorm(\mu_{z_{t}}, \sigma^{2}), & t &\ge 1,
\]
with $z_{1} \defined 1$. 
Let $\mcZ \defined (z_{t})_{t=1}^{T}$, $\mcY \defined (y_{t})_{t=1}^{T}$,
and $K \defined \max_{t=1,\dots,T} z_{t}$.
The joint density of the model is 
\[
\begin{split}
\lefteqn{\Pr(\mcZ, \mcY, \bmu_{1:K}, \bpi_{1:K,1:K}, \bbeta_{1:K} \given \sigma_{0}, \sigma, \alpha, \gamma)} \\
& = \Pr(\mcZ, \bbeta_{1:K}, \bpi_{1:K,1:K} \given \alpha, \gamma) 
\prod_{t=1}^{T} \distNorm(y_{t}\given \mu_{z_{t}}, \sigma^{2})
\prod_{k=1}^{K} \distNorm(\mu_{k}\given 0, \sigma_{0}^{2}),
\end{split}
\]
where 
\[
\begin{split}
\lefteqn{\Pr(\mcZ, \bbeta_{1:K}, \bpi_{1:K,1:K} \given \alpha, \gamma)} \\
&\defined 
\distNamed{GEM}(\bbeta_{1:K}; \gamma)
\prod_{k} \distDP(\bpi_{k,1:K}; \alpha\bbeta_{1:K})
\prod_{t=1}^{T} \distMulti(z_{t}; \bpi_{z_{t-1},1:K}). \label{eq:hdp-hmm-seq}
\end{split}
\]

We consider two approaches to obtaining the SVA for the HDP-HMM. 
The first is a ``combinatorial'' approach, in which we integrate
out $\bbeta_{1:K}$ and $\bpi_{1:K,1:K}$. 
The second is a ``direct'' approach, in which we do not integrate
out $\bbeta_{1:K}$ and $\bpi_{1:K,1:K}$. 

\subsection{Combinatorial Approach}

By integrating out $\bbeta_{1:K}$ and $\bpi_{1:K,1:K}$,
we obtain the CRF representation, which is the same urn scheme 
representation used in the original iHMM paper \citep{Beal:2002,Teh:2006b}.
The development is exactly the same as the HDP mixture model case,
see \cref{eq:hdp-crf,eq:log-hdp-crf-1,eq:log-hdp-crf-2}. 
As before $\beta = \frac{1}{2\sigma^{2}}$, $\gamma = \exp(-\lambda_{1}\beta)$, 
and $\alpha = \exp(-\lambda_{2}\beta)$, where $\lambda_{1}$
and $\lambda_{2}$ are free parameters: 
\[
\log\Pr(\bt, \mcZ \given \alpha, \gamma)
&= -\beta \lambda_{1} (K-1) - \beta \lambda_{2}\sum_{i=1}^{K} (t_{i\cdot} - 1) + O(1).
\]
If $s_{i}$ is the number of distinct transitions out of state $i$,
then 
\[
\max_{\bt \given \bt \sim \mcZ} - \lambda_{1} (K-1) - \lambda_{2}\sum_{i=1}^{K} (t_{i\cdot} - 1)
= - \lambda_{1} (K-1) - \lambda_{2}\sum_{i=1}^{K} (s_{i} - 1).
\]
If we use the free parameter $\lambda_3$ introduced in \cref{sec:SVAforHDP}, 
then the SVA optimization problem for the HDP-HMM is
\[
\min_{K,\mcZ,\bmu} \left\{
  \sum_{t=1}^{T}\|y_{t} - \mu_{z_{t}}\|^{2} + \lambda_{1} (K-1) 
+ \lambda_{2}\sum_{i=1}^{K} (s_{i} - 1) 
+ \lambda_{3}\sum_{i=1}^{K} \|\mu_{i}\|^{2} \right\}.  \label{eq:sva-hdp-hmm}
\]

In this cost function, the $\lambda_1$ term adds the cost for the total number of states
and $\lambda_2$ term penalizes the total number of distinct transitions out of the states.
As in \cref{eq:sva-hdp}, the last term represents the cost corresponding to the prior over the means of the states.

\subsection{Direct Approach}

Alternatively, we can choose not to integrate out 
$\bbeta_{1:K}$ and $\bpi_{1:K,1:K}$.
If we let
$\beta_{K+1} \defined 1 - \bar\beta_{K}$ and 
$\pi_{i,K+1} \defined 1 - \bar\pi_{i,K}$, then
\[
\begin{split}
\lefteqn{\Pr(\mcZ, \bbeta_{1:K}, \bpi_{1:K,1:K} \given \alpha, \gamma)} \\
&= \prod_{i=1}^{K} \distBeta\left(\frac{\beta_{i}}{1 - \bar\beta_{i-1}} \Big| 1, \gamma\right)
   \distDir(\bpi_{i,1:K+1} \given \alpha\bbeta_{1:K+1}) \prod_{t=2}^{T} \pi_{z_{t-1}z_{t}} \\
&= \prod_{i=1}^{K}\left\{\frac{\Gamma(1 + \gamma)}{\Gamma(\gamma)} \left(\frac{1 - \bar\beta_{i}}{1 - \bar\beta_{i-1}}\right)^{\gamma - 1} 
  \Gamma(\alpha) \prod_{j=1}^{K+1} \frac{\pi_{ij}^{\alpha\beta_{j}-1}}{\Gamma(\alpha\beta_{j})}\right\}
  \prod_{t=2}^{T} \pi_{z_{t-1}z_{t}}.
\end{split}
\]
Let $\gamma = \exp(-\lambda_{1}\beta)$ and $\alpha = \lambda_{2}\beta$.
Taking the logarithm of the product from $i=1$ to $K$ yields
\[
\begin{split}
&\sum_{i=1}^{K}\Bigg\{ -\log \Gamma(\gamma) + \alpha \log \alpha - \alpha + o(\beta) \\
&\phantom{\sum_{i=1}^{K}~\Bigg\{} + \sum_{j=1}^{K+1}\left\{ -\alpha\beta_{j} \log(\alpha\beta_{j}) + \alpha\beta_{j} + \alpha\beta_{j}\log \pi_{ij} + o(\beta) \right\}
\Bigg\} 
\end{split} \\
&= \beta \sum_{i=1}^{K}\left\{ -\lambda_{1} 
+ \sum_{j=1}^{K+1}\left\{ -\lambda_{2}\beta_{j} \log(\beta_{j}) + \lambda_{2}\beta_{j}\log \pi_{ij}\right\} 
\right\} + o(\beta) \\
&= -\beta\left\{\lambda_{1}K +  \lambda_{2}\sum_{i=1}^{K}\kl{\bbeta_{1:K+1}}{\bpi_{i,1:K+1}}\right\} + o(\beta).
\]
Here we have used the asymptotic expansions of $\log \Gamma(z)$:
\[
\log \Gamma(z) &=  z \log z - z + o(z), & z &\to \infty \\
\log \Gamma(z) &=  - \log z + O(z), & z &\downarrow 0.
\]
Hence, the SVA minimization problem for the HDP-HMM is:
\[
\begin{split}
\min_{K,\mcZ,\bbeta,\bpi}\Bigg\{&
 \sum_{t=1}^{T}\|y_{t} - \mu_{z_{t}}\|^{2} 
- \zeta\sum_{t=2}^{T} \log\pi_{z_{t-1}z_{t}} \\
&+ \lambda_{1}K 
+  \lambda_{2}\sum_{i=1}^{K}\kl{\bbeta_{1:K+1}}{\bpi_{i,1:K+1}} 
+ \lambda_{3}\sum_{i=1}^{K} \|\mu_{i}\|^{2}
\Bigg\}.
\end{split}
\]

Compared to \cref{eq:sva-hdp-hmm}, the main difference is in the terms involving 
hyperparameters $\zeta$ and $\lambda_2$. 
In this cost function, the $\zeta$ term penalizes transition 
probabilities very close to zero. 
The KL divergence term between $\bbeta_{1:K+1}$ and $\bpi_{i,1:K+1}$ is due to 
the hierarchical structure of the prior and it biases the transition 
probabilities $\bpi_{i,1:K+1}$ to be similar to the prior $\bbeta_{1:K+1}$.

\printbibliography

\end{document}